\documentclass{svproc}
\usepackage{hyperref}
\usepackage{natbib}
\usepackage{cleveref}
\usepackage{xcolor}
\usepackage{array}
\usepackage{adjustbox}
\usepackage{booktabs}

\begin{document}
\mainmatter              % start of a contribution
\title{A Comparative Analysis of Transformer
Models in Social Bot Detection}
\titlerunning{Social Bot Detection}  % abbreviated title (for running head)
\author{Rohan Veit \and Michael Lones}
\authorrunning{Veit et al.} % abbreviated author list (for running head)
\tocauthor{Rohan Veit and Michael Lones}
\institute{Department of Computer Science, Heriot-Watt University, Edinburgh, UK\\
\email{rv2009@hw.ac.uk}, \email{m.lones@hw.ac.uk}\\
}

\maketitle             % typeset the title of the contribution

\begin{abstract}
Social media has become a key medium of communication in today's society. This realisation has led to many parties employing artificial users (or bots) to mislead others into believing untruths or acting in a beneficial manner to such parties. Sophisticated text generation tools, such as large language models, have further exacerbated this issue. 
This paper aims to compare the effectiveness of bot detection models based on encoder and decoder transformers. Pipelines are developed to evaluate the performance of these classifiers, revealing that encoder-based classifiers demonstrate greater accuracy and robustness. However, decoder-based models showed greater adaptability through task-specific alignment, suggesting more potential for generalisation across different use cases in addition to superior observa. These findings contribute to the ongoing effort to prevent digital environments being manipulated while protecting the integrity of online discussion.
% We would like to encourage you to list your keywords within
% the abstract section using the \keywords{...} command.
\end{abstract}

\section{Introduction}
\label{sec:introduction}
As the internet has evolved, so have the services found on it; once benign platforms such as social media have transitioned from an opportunity for connecting with friends and self-expression to a tool used to spread information of varying credibility to the masses, shaping public beliefs and even influencing real-world events \cite{RN3}. Sites like X and Facebook allow individuals, organisations, or governments to communicate with unprecedented reach. This factor has cemented social media as a pillar of how today's narratives are formed, with them having a `profound influence on inter-personal communication' \cite{RN1}.

The influence of social media is particularly noticeable in its ability to amplify narratives. Posts that demonstrate polarizing and sensationalist views are promoted by algorithms favouring engagement rather than informed posts that display neutrality. One such example is the COVID-19 pandemic, where social media contributed to the dissemination of misleading information that led to vaccine avoidance and mask refusal \cite{FerreiraCaceresMariaMercedes2022Tiom}. Thus, it is clear that measures must be taken to prevent the pollution of digital spaces with fake information; identifying artificial users is an important component of this.

A `bot' is an artificial user of a platform that acts and appears as a real user would. When they first appeared, bots were generally easy to detect due to the repetitive nature of their broadcasted information, meaning they could be eliminated with rule-based detection (such as content-based thresholds like a duplicate message limit \cite{5741690}). However, modern bots have evolved to leverage advanced natural language processing tools, such as large language models, enabling them to better blend in with genuine users and evade existing detection measures \citep{feng2024doesbotsayopportunities}.

In this study, a variety of bot detection pipelines will be evaluated under different test cases to analyse their strengths, shortcomings and ideal applications.

%\textcolor{red}{Good enough summarising sentence?}

%

\section{Background}
\label{sec:background}
This section will briefly provide foundational context into bot detection within social media as well as the underlying technologies that will be investigated within this study. It will cover the justification for the selected case study and evaluate the findings of recent related studies and their shortcomings to highlight research gaps that should be investigated. Together these components will lay the groundwork for the methodology detailed in \cref{sec:methodology}.

%\textcolor{red}{Not sure how I feel about this intro}

\subsection{Bots}
\label{sec:back_terms_bots}

In \cite{ORABI2020102250}, a social media bot is defined as a computer algorithm that produces content and interacts with users. While this is not inherently sinister by definition (proven by helpful bots such as the now-defunct account \href{https://x.com/remindmetweets?lang=en}{\textit{RemindMeTweets}}, which allowed users to automatically get sent a reminder of a post after a specified timespan), this ability to coordinate hundreds or even thousands of accounts in unison can give way to malicious use such as spam campaigns \citep{10.1109/ICDM.2012.28}, artificial promotion of content \citep{spammersandpromoters} or even the manipulation of political events via a tactic named ``astroturfing''\citep{Ratkiewicz_Conover_Meiss_Goncalves_Flammini_Menczer_2021}.

\subsection{Transformer Models}

%\textcolor{red}{Is this necessary? I think it's good to provide context as to how (later mentioned) bot detection solutions literally function, but it feels a little out of place and obviously takes up a fair bit of room.}

Transformer models have become one of the frontier technologies for machine learning tasks. Unlike traditional feed-forward networks, rather than processing each input token sequentially, transformers use \textit{self-attention} to model the relation between tokens without the need for recurrent layers typically seen within previous models such as Long Short Term Memory (LSTM) networks. They can be categorised as encoder-only, decoder-only and encoder-decoder based on their architecture and how they generate output. Encoder-only transformers specialise in understanding and representing inputs through bidirectional self-attention, where each token attends (relates) to all other tokens in the input. These models excel in tasks that require a deep understanding of the input, such as text classification, sentiment analysis and feature extraction. Some examples of these transformers are BERT, RoBERTa and DistilBERT.

Decoder-only transformers use unidirectional self-attention, where each token can only attend to tokens that preceded it in the input. This makes them ideal for generative tasks where the model has to sequentially predict the next most likely word/token from a given prompt. However, they can be modified to complete other tasks through the process of prompt engineering (modifying the input provided to the model). Example use cases are text generation and completion. Well known models include \href{https://platform.openai.com/docs/models/o1}{OpenAI's GPT family} and \href{https://ai.google.dev/gemini-api/docs/models/gemini}{Google's Gemini}.

The encoder-decoder architecture combines both approaches by using an encoder to process the input and a decoder to generate the output. This makes it ideal for tasks requiring complex understanding and generation, such as translation or summarisation. Some examples of this approach are BART\citep{lewis2019bartdenoisingsequencetosequencepretraining} and T5\citep{raffel2023exploringlimitstransferlearning}.

\subsection{Datasets}
\label{sec:back_twitter}

Due to its large user base and scope, X (formerly Twitter) is a commonly used case study in bot detection experiments. We use the name ``Twitter`` from here on due to this being the naming convention followed in previous and recent academic research. Twitter is also one of the social media platforms most affected by the rise of bots, with 9--15\% of users being estimated to be bot accounts as of 2017 \citep{varol2017onlinehumanbotinteractionsdetection}. Its importance as a case study is further motivated by its use within large-scale bot attacks, for example for political interference \citep{Ratkiewicz_Conover_Meiss_Goncalves_Flammini_Menczer_2021}.

TwiBot-22 \cite{feng2023twibot22graphbasedtwitterbot} is a dataset that builds upon predecessors (Cresci-2015 \cite{CRESCI201556}, Varol-2017 \cite{varol2017onlinehumanbotinteractionsdetection}, TwiBot-20 \cite{Feng_2021}). It consists of 1,000,000 users, making it nearly 5 times larger than the next largest social media bot dataset, TwiBot-20, which contains 229,000 users. TwiBot-22 also uses a distribution diversity biased algorithm to capture a wide demographic of users, while  maintaining a full graph structure so that relations between users present in the dataset are also captured. Labels for users are then generated using a combination of 8 hand-crafted labels and 7 competitive feature-based classifiers which are then refined using Snorkel \cite{Ratner_2017} to generate the final data annotations. These labels were found to be 10.5\% more accurate than TwiBot-20.

%\textcolor{red}{More here about how the subset of data used in method was obtained? Or is this better placed in the method section?}

\subsection{Transformer Models to Detect Bots}
\label{sec:back_transformer}

%\textcolor{red}{Maybe further cut this down by omitting first example? Last two are good demonstrations of an encoder study + decoder study, but lack clear comparison. I do think that this is the most important part of the background though}

Various studies have attempted to use transformers to distinguish social bots from genuine users in social networks.

In \cite{9385071}, transformers were used to create a robust and language-agnostic bot detection system, employing both BERT and RoBERTa to encode text features from user accounts. They used multilingual base versions of BERT and RoBERTa to generate 768 and 1024-dimensional embedded vector representations of the input features (such as usernames, descriptions, and language indications) and then concatenated these with additional user meta-data for the final classification carried out by a custom deep neural network named Bot-DenseNet, opting to use Scaled Linear Exponential Linear Unit (SELU) as an activation function over the traditional ReLU function. To address the unbalanced nature of the provided dataset, stratified sampling was utilised. 

This approach was later refined using a variety of transformer-based approaches. In \cite{10630818}, pre-trained language models (PLMs)---including various BERT models and the encoder-only model GPT-3---were fine-tuned on bot-specific datasets (TweepFake \cite{Fagni_2021} and \href{https://github.com/osome-iu/AIBot_fox8}{fox8-23}) which then fed into a classical feed-forward neural network that used outputs from these PLMs to classify users. These models were able to consider contextual nuances in social media text---consisting of number frequency, user mentions, hashtags, URLs, emojis and topic analysis---through their dynamic embedding capabilities, offering greater flexibility and accuracy over static embedding methods such as GloVe \citep{pennington2014glove} and Word2Vec \citep{mikolov2013efficientestimationwordrepresentations}.

Decoder-only transformers have also been demonstrated to be of benefit to honing the bot classification task. A study detailed a solution wherein three Large Language Models (LLMs) that were 'instruction tuned' on 1,000 labelled examples. This consisted of providing the LLMs with the standard user metadata and textual data used in previous solutions, as well as a subset of the user's neighbourhood. Prompts provided to the LLMs were then tuned to provide in-context examples of both classes. This method outperformed existing bot-detection measures by 9.1\% \citep{feng2024doesbotsayopportunities}.
Furthermore, the LLM displayed a reasonable understanding of the graph structure inherent in social networks, with the LLM being able to suggest additions and removals from the following list of a user to make it less vulnerable to bot detection methods. In conclusion, this highlights that although decoder-transformers have a lot to offer towards bot detection, their incremental sophistication also marks a double-edged sword as these advances could be utilised by malicious actors.

Previous work shows that transformers have a lot of potential for bot detection. However, there is little understanding of the link between architecture and capability. In particular, whilst the advent of powerful decoder-only models such as GPT-4 has brought interest in using these for bot detection, there has been little research into how well these work in comparison to more established encoder-based models. In this work, we address this gap. %In \cref{sec:methodology}, we will detail the method for conducting a thorough analysis into the effectiveness of decoder versus encoder-based solutions to social bot detection.

\section{Methodology}
\label{sec:methodology}
%\textcolor{red}{WIP: sort of unsure how to concisely describe each pipeline and map them to the experiments carried out in \cref{sec:results}. It sort of feels that the experiments appear out of nowhere with no justification as to why they were selected.}

\subsection{Data Preprocessing}

\begin{figure}
    \centering
    \includegraphics[width=0.8\linewidth]{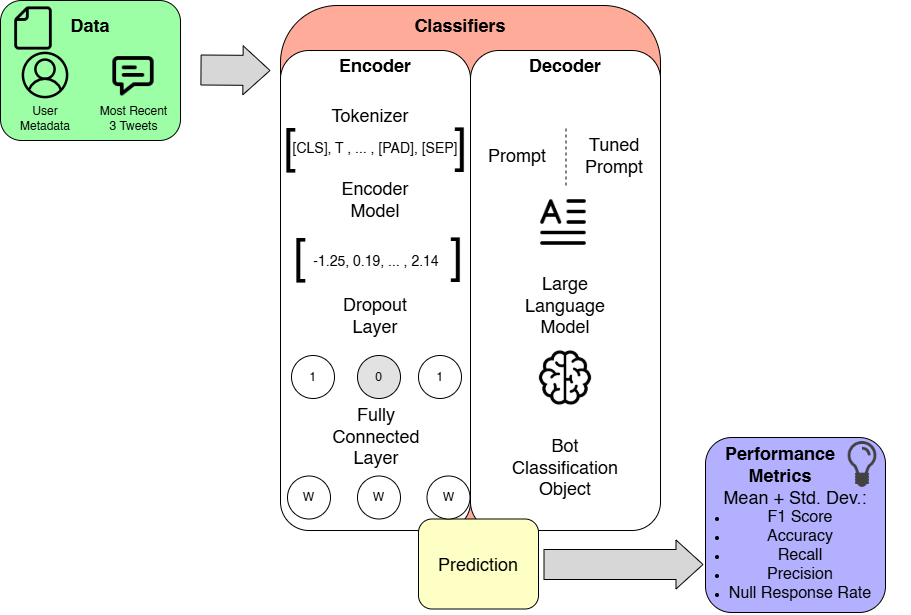}
    \caption{Method Summary Diagram}
    \label{fig:method_summ}
\end{figure}

Firstly, data was obtained from the TwiBot-22 dataset. Due to limitations in the computational resources of the development environment used, only the first 10\% of user messages were extracted to maintain a dataset of workable size. From this, 5 train/test partitions of 1,000 users were derived to be used in five-fold cross-validation.

With the data cleaned, it was then necessary to define the structured data types that would be provided to the models for inference. For this, the Python library Pydantic was utilised for data schema definition and robust validation of data. Firstly, data schemas for users and tweets were defined from the cleaned JSON data. From these, corresponding Python objects could easily be instantiated, with rigorous type validation to ensure the handling of malformed or missing fields.

\subsection{Decoder Pipeline Creation}

Following data preprocessing, the pipelines used for evaluating the decoder models were architected. The pipeline was designed to facilitate structured data processing, model inference, and ultimately a binary predication of whether the inputted user was likely to be an artificial account or not.

Ollama was selected for implementing model inference. Ollama is an LLM inference framework designed to facilitate inference with LLMs in a local or cloud-based environment, providing a simple and efficient way to run models on a user's own hardware. This results in reduced latency and dependence on external API services. This means that users wishing to deploy this or a similar pipeline can be assured of model uptime and have increased control over model type and configuration. It is worth noting that Ollama only supports a select library of LLMs and due to the requirement of the pipeline having structured data as an input and output (to facilitate utilisation within real-world contexts), this limited the selection of LLMs further to those that supported tool usage.
The selected LLMs for analysis were chosen to be: LLama3.1, Mistral, and Qwen2.5.

To facilitate easy implementation within real-world contexts, the choice was made for the pipeline to provide responses in the form of structured objects. This meant the employment of PydanticAI, a Python framework that builds on the data class \texttt{BaseModel} interface provided in Pydantic to allow the creation of agents that enforce strict input and output formats defined by data schemas. This aids the development of applications require reliable outputs, common in systems that use automated decision-making, such as social bot detection. The resultant PydanticAI agent was defined to take a prompt in the format of a string paragraph (to support prompt engineering) as input, with a defined 'bot detection response' data class as an output: an object that consisted of a boolean 'is\_bot' indicator accompanied by a reason for the selected choice to provide model reasoning observability and to better guide models towards reasonable responses.

\begin{figure}[h!]
    \centering
    \includegraphics[width=\linewidth]{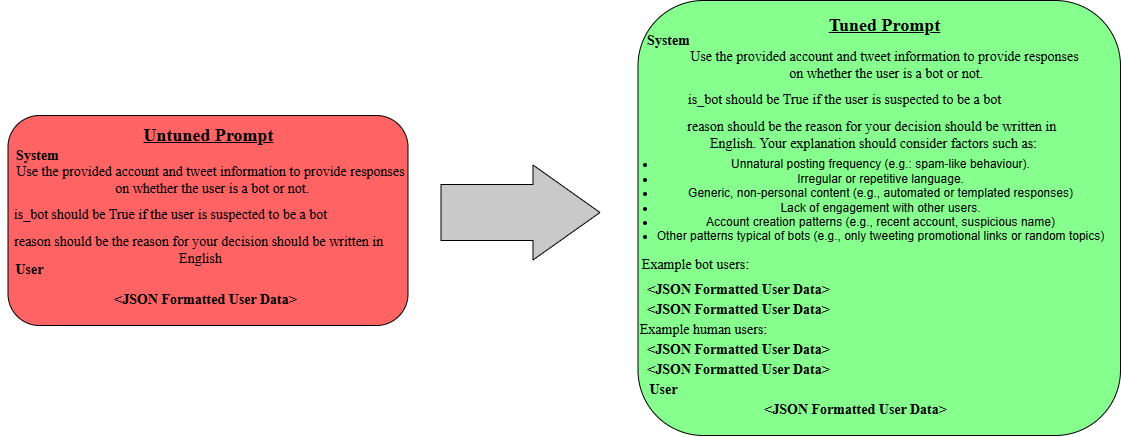}
    \caption{Untuned and Tuned Decoder Prompts}
    \label{fig:prompts}
\end{figure}

Finally, to facilitate a comparative analysis of the classifier's responsiveness to task-specific alignment, prompt engineering was employed. Specifically, few-shot learning was employed to provide the model example users and their labels to help steer the model towards extrapolating this logic to unseen examples. To do this, four example users from outside the five test splits were extracted and inserted into the model's prompt within a 'Example bot/human users' section. Finally, task specification was also used to help aid the model towards making decisions based on specific user features (i.e., low amount of posts or recent account creation).

Following the creation of the data and decoder pipelines, the development of the encoder pipeline began. This pipeline was responsible for transforming the raw textual data into a format compatible with an encoder model, the output from which could be used to classify whether the inputted data was likely to belong to an artificial user.

\subsection{Encoder Pipeline Creation}

The encoder pipeline was developed using HuggingFace's Transformers library and PyTorch, two widely adopted frameworks used for implementing machine learning models. Firstly - similar to the creation of the decoder pipeline - the internal encoder model selections had to be made. Considering previous literature, the resultant models selected for analysis were: DistilBERT, BERT, and RoBERTa.

With the internal models selected, work began on designing the classifier; firstly, the encoder part of the architecture was implemented using HuggingFace's \texttt{BertModel} interface which allows for models to easily be imported from the \href{https://huggingface.co/models}{model repository} also maintained by HuggingFace simply by changing the provided model name. Then, also using the same interface, forward passes could easily be conducted on the provided model to generate semantic embeddings representative of the inputted textual user data. The embeddings were then fed into a dropout layer to regularise the output data and prevent model overfitting, ensuring that the encoder models did not have an advantage over the decoder models due to their requirement of a subset of data used for training. Funally, the dropout layer fed into a fully-connected linear layer that was responsible for mapping the embedding representations to a final decision in the form of a confidence score for either class (bot or human).

With the data pipeline already implemented, it could readily be imported into the encoder experiment. Then, HuggingFace was used again to acquire a tokenizer (specific to the encoder model selected for that experiment) that would be used to create a PyTorch dataset class that would be responsible for handling the batching and shuffling of data.

Unlike the LLM component in the decoder pipeline which comes pre-trained, the neural network component of the encoder pipeline requires this phase to determine which features of a provided input example are significant for classification. To train the encoder classifier, the model was provided an example to provide a forward pass and make a prediction on. Then, using PyTorch's \texttt{CrossEntropyLoss} criterion, the loss of the model's prediction is calculated and then used for backpropagation using PyTorch's autograd system. This computes gradients of the loss with respect to the model's parameters that were then used by an optimiser using the Adam algorithm with weight decay fix to update model parameters. 

Finally, similarly to the decoder pipeline, model finetuning was conducted. This consisted of forming a list of hyper parameters to be modified, as well as values for which they would be analysed. Selected hyperparameters for finetuning were: Training epochs, alternate optimisers, and learning rate. Using these, an exhaustive grid search was conducted to find the best-performing combination of hyperparameter values that would be used for the fine-tuned comparative analysis (detailed in \cref{sec:res_experiment3}).

\section{Results and Observations}
\label{sec:results}
This section presents the findings of the experiments outlined in \cref{sec:methodology}. Each subsection details the performance of the decoder and encoder classifiers under different conditions, evaluating their performance via accuracy, precision, recall, F-1 score and null response rate. Additionally, standard deviations of these metrics have been provided to assess model variability.

\subsection{Experiment 1: Partial Feature Set}
\label{sec:res_experiment1}

\begin{table}[tb!]
    \begin{adjustbox}{width=\textwidth,center}
    \begin{tabular}{l|c|c|c|c|c}
    \hline
        \textbf{Model} & \textbf{Accuracy} & \textbf{Precision} & \textbf{Recall} & \textbf{F1-Score} & \centering\arraybackslash \textbf{Null Response Rate}  \\ \hline
        \multicolumn{6}{c}{\textbf{Decoder Models}} \\ \hline
        Llama3.1:8b & 0.505 ± 0.014 & 0.530 ± 0.014 & \textbf{0.606} ± 0.016 & 0.566 ± 0.015 & 0.051 ± 0.004 \\
        Mistral:7b & 0.299 ± 0.006 & \textbf{0.691} ± 0.053 & 0.177 ± 0.027 & 0.281 ± 0.036 & 0.453 ± 0.006 \\
        Qwen2.5:7b & \textbf{0.569} ± 0.020 & 0.585 ± 0.020 & 0.580 ± 0.037 & \textbf{0.582} ± 0.028 & \textbf{0.023} ± 0.006 \\ \hline
        \multicolumn{6}{c}{\textbf{Encoder Models}} \\ \hline
        BERT & 0.759 ± 0.017 & 0.766 ± 0.029 & 0.749 ± 0.035 & 0.757 ± 0.016 & \textbf{0.000} ± 0.000 \\
        DistilBERT & 0.759 ± 0.015 & \textbf{0.767} ± 0.022 & 0.744 ± 0.027 & 0.755 ± 0.015 & \textbf{0.000} ± 0.000 \\
        RoBERTa & \textbf{0.770} ± 0.008 & 0.757 ± 0.011 & \textbf{0.796} ± 0.023 & \textbf{0.776} ± 0.009 & \textbf{0.000} ± 0.000 \\ \hline
    \end{tabular}
    \end{adjustbox}
    \caption{Experiment 1 Results (Mean ± Standard Deviation)}
    \label{tab:exp1_results}
\end{table}

This experiment established a baseline for classifier performance by evaluating both the decoder and encoder models using a limited feature set consisting of only account metadata. This experiment aimed to assess the initial classification ability displayed by the two model architectures and serve as a reference point for further experiments.

The results are shown in \cref{tab:exp1_results} with the best-performing classifier for each metric of each architecture emboldened. % Alternatively, find the graphical form in \cref{}.
%\subsubsection{Results}
This shows that the encoder models outperformed the decoder models in all metrics. This may be due to the explicit training of the downstream classifier in the encoder models, providing better alignment to the task domain. This shows that, by utilising supervised learning,  performance can be increased by a significant margin. However, the results do also demonstrate the ability of the decoder models to operate to some degree without any additional training.
% means that it will likely be better at generalising and therefore may be more eligible for being applied to other tasks with minimal engineering.

All models also demonstrated a low standard deviation, demonstrating consistency across training runs.
However, the decoder models noticeably produced some null responses, i.e.~they did not output class labels for all samples in the test set. This suggests they may be less suitable for real-world deployment where a binary response is required.
%The results clearly show the encoder classifiers to outperform the decoders, at least in the initial tests.
%Furthermore, a comparison of the two architectures' null response rate also shows the encoder architectures to be much more robust, indicating that encoder architectures may be more suitable for real-world employment where a binary response is required.

\subsection{Experiment 2A: Enriched Feature Set}
\label{sec:res_experiment2a}

This experiment aims to evaluate model performance under an enriched feature set. The primary objective of this experiment is to evaluate how a change in input features affects the performance metrics of each model, identifying which may be sensitive to feature modifications.
To systematically measure these effects, \cref{tab:exp2a_results} compares the results from Experiment 2A against those from Experiment 1, highlighting relative gains or losses in performance metrics.

%\subsubsection{Results}

\begin{table}[tb!]
    \begin{adjustbox}{width=\textwidth,center}
    \begin{tabular}{l|c|c|c|c|c}
    \hline
        \textbf{Model} & \textbf{Accuracy $\Delta$} & \textbf{Precision $\Delta$} & \textbf{Recall $\Delta$} & \textbf{F1-Score $\Delta$} & \textbf{Null Response Rate $\Delta$}  \\ \hline
        \multicolumn{6}{c}{\textbf{Decoder Models}} \\ \hline
        Llama3.1:8b & \textbf{-0.040} & \textbf{+0.001} & \textbf{+0.036} & \textbf{+0.016} & +0.081 \\
        Mistral:7b & -0.053 & -0.072 & -0.060 & -0.0854 & \textbf{+0.079} \\
        Qwen2.5:7b & -0.156 & -0.010 & -0.131 &-0.079 & +0.228 \\ \hline
        \multicolumn{6}{c}{\textbf{Encoder Models}} \\ \hline
        BERT & \textbf{+0.000} & -0.011 & \textbf{+0.018} & \textbf{+0.004} & \textbf{+0.000} \\
        DistilBERT & -0.004 & -0.016 & +0.020 & +0.002 & \textbf{+0.000} \\
        RoBERTa & -0.015 & \textbf{-0.008} & -0.028 & -0.017 & \textbf{+0.000} \\ \hline
    \end{tabular}
    \end{adjustbox}
    \caption{Experiment 2A Results (Mean Difference From Experiment 1)}
    \label{tab:exp2a_results}
\end{table}

%The results from experiment 2A show that
In general, the decoder models decreased in performance when using the full feature set. On average, accuracy dropped across all the decoder models, with the most significant decline being observed in Qwen2.5. This model also showed considerable reductions in recall and F1-Score. Furthermore, all the decoder models experienced an increase in null-response rate, indicating that a fuller feature set affects classifier robustness.
The encoder models displayed more stable performance, with relatively minor changes in most of the metrics. BERT maintained its accuracy while improving in three out of the four recorded metrics. RoBERTa showed a drop across all recorded metrics.

%\subsubsection{Analysis}

%From the results of experiment 2A, it can be seen that both architectures of classifier generally responded poorly to the enhanced featureset.
The reduced performance indicates that additional text data (in the form of the user's most recent three tweets) is counterproductive, adding noise to the process of determining features that contribute to labels. Though perhaps surprising, this observation is in line with other studies, such as \cite{feng2024doesbotsayopportunities} where adding extra text to user metadata reduced performance across Mistral, Llama2, and ChatGPT. However, encoder models showed lower sensitivity in our results.
%, aside from one instance where additional prompt-engineering was conducted on ChatGPT's prompt.

%It should be noted that although both classifier architectures demonstrated decreased performance, the encoder models generally showed less of a decrease, implying that they may respond better to larger noisier feature sets.

\subsection{Experiment 2B: Model Size Comparison}
\label{sec:res_experiment2b}

\begin{table}[tb!]
    \centering
    \begin{adjustbox}{width=\textwidth,center}
    \begin{tabular}{l|c|c|c|c|c}
    \hline
        \textbf{Model} & \textbf{Accuracy} & \textbf{Precision } & \textbf{Recall} & \textbf{F1-Score} & \textbf{Null Response Rate}  \\ \hline
        Qwen2.5 3b & 0.520 ± 0.015 & 0.594 ± 0.037 & 0.255 ± 0.018 & 0.357 ± 0.024 & 0.040 ± 0.003 \\
        Qwen2.5 7b & 0.413 ± 0.019 & 0.573 ± 0.027 & \textbf{0.448} ± 0.023 & \textbf{0.503} ± 0.025 & 0.251 ± 0.017 \\
        Qwen2.5 14b & \textbf{0.558} ± 0.014 & \textbf{0.660} ± 0.034 & 0.246 ± 0.019 & 0.358 ± 0.024 & \textbf{0.002} ± 0.001 \\
        \hline
    \end{tabular}
    \end{adjustbox}
    \caption{Experiment 3 Results (Mean ± Standard Deviation)}
    \label{tab:exp2b_results}
\end{table}

This experiment investigates the role of model size by repeating experiment 2A with 3, 7 and 14 billion parameters variants of Qwen2.5, whose 7 billion parameter version was the strongest performing encoder model in experiment 1. The results are shown in \cref{tab:exp2b_results}. 
%In this experiment, models were provided with the same expanded feature set as in experiment 2A. The aim was to determine the relationship between decoder model parameter count and bot classification performance. Find the results detailed in \cref{tab:exp2b_results}.
%\subsubsection{Results}
This shows that the larger 14 billion parameter model improves in accuracy and null response rate in comparison to the 7 billion parameter model. However, its performance is still lower than when the reduced feature set is used with the 7 billion parameter model, suggesting that more parameters does not enable it to benefit from the extra information provided.

Additionally, recall and F1-score are reduced for the larger model. From inspecting the 7 billion parameter model's confusion matrix, this is because the intermediate model seems to have a stronger bias towards predicting `bot' than both the 3 billion and 14 billion variants, with it predicting `bot' 40\% of the time compared to Qwen2.5:14b which only selected `bot' 18\% of the time. This may be due to the 7 billion parameter model being the `default' release of Qwen2.5, suggesting it may have undergone further optimisation.

%\subsubsection{Analysis}
%As expected, the largest model of 14 billion parameters generally performed the best during this experiment, with the positive trend between the 3 billion and 14 billion parameter models looking as expected. However, as stated above, the 7 billion parameter variant performed the best when classifying 'bots' due to its higher classification proportion of 'bot' predictions (40\% compared to the 14 billion's 22\%). 

\subsection{Experiment 3: Tuned Classifiers}
\label{sec:res_experiment3}

The objective of this experiment was to compare how the different architectures responded to task-specific tuning. The selected encode and decoder models were those that performed the best during experiment 2. Both models were provided the same feature set from experiment 2 to ensure a fair comparison could be made. To tune the encoder classifier, hyperparameter fine-tuning was employed, and for the decoder classifier, prompt engineering was adopted.

%\subsubsection{Results}

\begin{table}[tb!]
    \centering
    \begin{adjustbox}{width=\textwidth,center}
    \begin{tabular}{l|c|c|c|c|c}
    \hline
        \textbf{Model} & \textbf{Accuracy} & \textbf{Precision } & \textbf{Recall} & \textbf{F1-Score} & \textbf{Null Response Rate}  \\ \hline
        Qwen2.5:14b & 0.655 ± 0.022 & 0.670 ± 0.023 & 0.617 ± 0.025 & 0.642 ± 0.023 & 0.001 ± 0.001 \\
        RoBERTa & \textbf{0.774} ± 0.004 & \textbf{0.763} ± 0.012 & \textbf{0.794} ± 0.026 & \textbf{0.778} ± 0.019 & \textbf{0.000} ± 0.000 \\ \hline
    \end{tabular}
    \end{adjustbox}
    \caption{Experiment 4 Results (Mean ± Standard Deviation)}
    \label{tab:exp3_results}
\end{table}

The results are shown in \cref{tab:exp3_results}. As this shows, the decoder classifier was much more receptive to tuning, increasing its accuracy by 9.7\% from experiment 2B, suggesting that prompt engineering is beneficial within this context. The encoder classifier still outperformed it though, although only increasing in performance from experiment 2A comparatively marginally. Nevertheless, it achieved the highest mean accuracy, precision and F1 scores across all experiments.
%\subsubsection{Analysis}
%The results from experiment 3 highlight each classifier's differing responsiveness to task-specific tuning. Notably, the decoder classifier significantly improved, suggesting that prompt engineering improved the model's performance for this task. 
Another notable observation is that after tuning, the null response rate dropped to the lowest across all experiments, indicating that task-specific tuning also aids robustness. It should be noted that this tuning method requires no modification of the model's parameters, demonstrating the increased accessibility to task-specific tuning that decoder models provide.

%Despite this, the encoder classifier outperformed Qwen2.5:14b across all recorded metrics, achieving the highest accuracy, precision and F-1 score across all experiments.

%

\section{Limitations}

Due to the availability of data, this work was only evaluated on one dataset, TwiBot-22. This may limit the generality of the conclusions. Furthermore, while an analysis into a reduced and enhanced feature set was made, the full information captured by TwiBot-22 (such as user account relations) was not utilised. Finally, due to both financial and time cost, only three open-source models of each architecture were evaluated, with model size comparison only being carried out on sizes that were within the constraints of the experimental environment.

\section{Conclusions}
\label{sec:conclusion}

%\subsection{Summary}
This study conducts an in-depth analysis of the effectiveness of decoder and encoder-based social bot detection models, employing a variety of model types, sizes, and feature sets. %Through a series of experiments, the study assesses classifiers' performance and robustness to determine the effectiveness of each solution and how transferable it may be to real-world contexts.
Our results indicate that encoder-based architectures consistently outperform decoder-based models across all evaluation metrics, indicating that for bot detection tasks where supervised learning can be implemented, they may offer superior performance. Decoder-based approaches showed less promising results, but larger potential for improvement due to the ease with which the underlying model can be substituted.
%Furthermore, the decoder models show a positive relationship between internal model complexity and performance, highlighting this point. 
Decoder-based classifiers also showed greater responsiveness to task-specific tuning methods through prompt engineering, demonstrating their potential for low-cost improvement.

In future work, we plan to: evaluate larger, more complex models; evaluate a more diverse set of encoder (potentially outside of the BERT family, such as T5) and decoder models; and engineer a fuller feature set that captures more characteristics of the user's social network. 
%\begin{itemize}
%    \item Use the provided pipelines for evaluating larger, more complex models.
%    \item Use the provided pipelines for evaluating a more diverse set of encoder models (potentially outside of the BERT family, such as T5).
%    \item Use the provided pipelines for evaluating a more diverse set of decoder models.
%    \item Engineering a fuller feature set that captures more of the related social network.
%\end{itemize}

\bibliographystyle{abbrv}
\bibliography{references}

\end{document}